\pdfoutput=1

\documentclass[10pt,twocolumn,letterpaper]{article}

\usepackage{iccv}
\usepackage{times}
\usepackage{epsfig}
\usepackage{graphicx}
\usepackage{amsmath}
\usepackage{amssymb}
\usepackage{enumerate}
\usepackage{multirow}
\usepackage{adjustbox}
\usepackage{authblk}

\def\be{ \begin{equation} }
\def\ee{ \end{equation} }
\def\bea{ \begin{eqnarray} }
\def\eea{ \end{eqnarray} }

\usepackage[pagebackref=true,breaklinks=true,letterpaper=true,colorlinks,bookmarks=false]{hyperref}

\iccvfinalcopy 

\def\iccvPaperID{****} 

\begin{document}

\title{Efficient and Scalable View Generation from a Single Image using Fully Convolutional Networks}

\author[1]{Sung-Ho Bae}
\author[2]{Mohamed Elgharib}
\author[3]{Mohamed Hefeeda}
\author[1]{Wojciech Matusik} 

\affil[1]{MIT CSAIL}
\affil[2]{QCRI, Hamad Bin Khalifa University}
\affil[3]{Simon Fraser University}

\affil[ ]{\tt\small \{sunghbae@mit.edu, melgharib@qf.org.qa, mhefeeda@sfu.ca, wojciech@csail.mit.edu\}}

\maketitle

\begin{abstract}
   Single-image-based view generation (SIVG) is important for producing 3D stereoscopic content. Here, handling different spatial resolutions as input and optimizing both reconstruction accuracy and processing speed is desirable. Latest approaches are based on convolutional neural network (CNN), and they generate promising results. However, their use of fully connected layers as well as pre-trained VGG forces a compromise between reconstruction accuracy and processing speed. In addition, this approach is limited to the use of a specific spatial resolution. To remedy these problems, we propose exploiting fully convolutional networks (FCN) for SIVG. We present two FCN architectures for SIVG. The first one is based on combination of an FCN and a view-rendering network called DeepView$_{ren}$. The second one consists of decoupled networks for luminance and chrominance signals, denoted by DeepView$_{dec}$. To train our solutions we present a large dataset of 2M stereoscopic images. Results show that both of our architectures improve accuracy and speed over the state of the art. DeepView$_{ren}$ generates competitive accuracy to the state of the art, however, with the fastest processing speed of all. That is x5 times faster speed and x24 times lower memory consumption compared to the state of the art. DeepView$_{dec}$ has much higher accuracy, but with x2.5 times faster speed and x12 times lower memory consumption. We evaluated our approach with both objective and subjective studies.
\end{abstract}

\section{Introduction}
Single-image-based view-generation (SIVG) is a technique to generate a new view image from a single image. It has typically aimed to generate the right-view image from a left-view image in stereoscopic viewing scenario \cite{xie2016deep3d}. SIVG has been actively studied over decades \cite{xie2016deep3d, zhuo2009recovery, hoiem2005automatic, cozman1997depth, saxena2009make3d, chen2016single, cozman1997depth, appia2014fully, baig2014im2depth, eigen2014depth, liu2015deep, calagari2015gradient, kim20102d, konrad2013learning, karsch2014depth}, as it can be widely applied to provide richer representations and important perceptual cues on image understanding as well as 3D modeling \cite{silberman2012indoor, saxena2009make3d}, etc. Especially, SIVG techniques are becoming more and more important, as the dominant portions of the 3D movie and virtual reality (VR) markets are coming from 3D content production and consumption \cite{matusik20043d, avila2014virtual}. Furthermore, computationally efficient SIVG methods can bring significant benefits as many 3D contents are consumed on mobile devices. In this paper, we focus on both computationally efficient and accurate SIVG methods.

SIVG consists of two main stages: single-image-based depth estimation (SIDE) and depth image-based-rendering (DIBR) \cite{xie2016deep3d}. Mathematically, SIDE and DIBR can be formulated as
\be
\hat{D}=h(L|\Theta_{h}),
\label{eq_1}
\ee
\be
\hat{R} =g(L, \hat{D}|\Theta_{g})
\label{eq_2}
\ee
where $L$, $\hat{D}$, $\hat{R}$ are the left-view image, estimated depth map, and estimated right-view image, respectively. $h(\cdot )$ in Eq. 1 and $g(\cdot )$ in Eq. 2 are considered SIDE and DIBR, respectively. $\Theta_{h}$ in Eq. 1 and $\Theta_{g}$ in Eq. 2 are model parameters. This paper uses disparity and depth interchangeably based on an assumption of the standard stereoscopic viewing condition where depth and disparity values are linearly proportional \cite{kalantari2016learning}. 

Most existing approaches have focused on estimating accurate depth maps, that is, they aimed to model $h(\cdot)$ in Eq. 1. Previous depth estimation approaches have relied on utilizing one or a few depth cues in images, such as color \cite{konrad2013learning}, scattering \cite{cozman1997depth}, defocus \cite{zhuo2009recovery}, and saliency regions \cite{kim20102d}. A recent breakthrough in depth estimation was achieved by convolutional neural network (CNN)-based data-driven approaches \cite{liu2015deep, eigen2014depth, chen2016single}. Although these approaches showed remarkable performance improvement compared to previous hand-craft features-based approaches, DIBR in Eq. 2 should additionally be performed for SIVG, which often involves much computation and visual distortion \cite{xie2016deep3d}. 

Recently, Xie \etal~\cite{xie2016deep3d} proposed an end-to-end SIVG method, called Deep3D. The network architecture of Deep3D relies on a pre-trained CNN (VGG-16 Net \cite{simonyan2014very}) with a rendering network. Deep3D showed the highest prediction performance in SIVG compared with the CNN-based depth-estimation methods followed by a standard DIBR \cite{zhuo2009recovery}. However, Deep3D requires large memory space with much computational complexity as it relies on pre-trained CNNs with fully-connected (dense) layers. Also, the dense layers in Deep3D inevitably limit the spatial resolution of its input and output (\ie, 384$\times$160), thus constraining flexibility and applicability. In order to remedy the aforementioned problems, we propose to exploit a fully-convolutional-network (FCN) for SIVG.  
Our work is inspired by the recent success of FCNs in super-resolution \cite{kim2015accurate, long2015fully, mao2016image}.

Aspects of novelty in our work include:
\begin{enumerate}
\item We propose a new network for efficient SIVG by combining an FCN with a rendering network. We call this DeepView$_{ren}$. Thanks to our simple and efficient FCN architecture, DeepView$_{ren}$ runs x5 times faster than the state of the art \cite{xie2016deep3d}, with x24 times lower memory consumption. It also achieves competitive prediction accuracy.
\vspace{-0.2cm}
\item We present a decoupled architecture for luminance and chrominance signals, denoted by DeepView$_{dec}$. Here, two networks train and infer the Y and CbCr signals separately. This shows much higher prediction performance than the state of the art \cite{xie2016deep3d}. However, with only x2.5 times faster and x12 times lower memory consumption.  
\vspace{-0.2cm}
\item Thanks to exploiting an FCN, our methods can take input of various-sized images and outputs correspondingly-sized ones. This spatial scalability was not present in existing techniques, mainly due to their dense layers.
\vspace{-0.2cm} 
\item We collected a very large dataset of 27 non-animated  stereoscopic movies having a total of 2M frames. To the best of our knowledge, there are no sufficiently large publicly available datasets for training SIVG. We are planning to release all our code and data to encourage future research. 
\end{enumerate}

This paper is organized as follows: we review related work and introduce our architectures in Section 2 and 3, respectively; we perform thorough experiments to explore efficient network architecture with spatial scalability; we compare the proposed method with the state-of-the-art SIVG method in Section 5; Section 6 concludes our work.

\section{Related Work}

Our work exploits the advances of FCNs in super-resolution and rendering network in view generation. 

\vspace{0.1cm}
\noindent
{\bf FCN-based super resolution} Our architecture is inspired by recent success of FCN in super resolution problems  \cite{kim2015accurate, kim2015deeply, dong2016image, mao2016image}. Dong \etal~ proposed a three layered FCN and showed powerful performance in both accuracy and efficiency. Kim \etal~ further extended the work in \cite{dong2016image} by establishing a very deep (20 layered) FCN architecture \cite{kim2015accurate}. To boost the network convergence, they adopted a residual learning and gradient clipping method \cite{kim2015accurate}. Very recently, Mao \etal~ proposed a symmetric skip connection between encoding and decoding networks to transfer high frequency details to the output \cite{mao2016image}. All results in \cite{dong2016image,kim2015deeply, mao2016image} reveal that FCNs can be effectively applied for per-pixel regression problems with much efficiency and accuracy.

\vspace{0.1cm}
\noindent
{\bf Monocular depth estimation} Recently, CNN-based monocular (single image) depth prediction methods have shown promising performance in accuracy compared to previous hand-craft features-based approaches \cite{liu2015deep, eigen2014depth, chen2016single}. Eigen \etal~ proposed a multi-scale CNN to secure large receptive field sizes \cite{eigen2014depth}. Here, the receptive field size in CNN corresponds to the range of contextual information used for inference. Results show securing large-receptive fields in a network is important for monocular depth prediction since it reduces uncertainty of depth relations between different objects. Liu \etal~ incorporate a CRF learning scheme into CNN in order to estimate accurate depth maps from a single image \cite{liu2015deep}. To make the CRF-learning within CNN tractable, they derived a closed form solution. Chen \etal~ introduced a new dataset called Depth In the Wild (DIW), consisting of relative depth points taken in unconstrained settings \cite{chen2016single}. They trained an inception-like-network to generate a relative depth metric. This widens the applicability of monocular depth prediction methods. Although existing SIDE methods show good accuracy in mapping an image to depth, the estimated depth-maps are intermediate representations in SIVG and and hence still require a DIBR process. Such process is often computationally expensive and prone to errors. Compared to existing depth-estimation methods, our approach is an end-to-end mapping that efficiently combines SIDE and DIBR into one network. Hence, we do not need a separate DIBR block. 

\vspace{0.1cm}
\noindent
{\bf View generation} Recently, Flynn \etal~ proposed DeepStereo that takes a set of calibrated images as input and outputs images of new views \cite{flynn2015deepstereo}. DeepStereo consists of a rendering network and a color image generation network. The rendering network generates probabilistic disparity maps and renders a new view by multiplying the disparity maps with outputs of the color image generation network. Very recently, Xie \etal~ proposed an end-to-end SIVG method \cite{xie2016deep3d} based on a pre-trained CNN (VGG-16 Net \cite{simonyan2014very}) and the rendering network. To estimate a right-image from a single left-image, Deep3D extracts features from the pre-trained network. The extracted features are up-scaled with deconvolution layers (\ie, convolution layers with strides over 1) and are directly fed into the rendering network. Finally, the rendering network generates a right-view image. 


Fully connected (dense) layers in pre-trained networks have limited input/output dimensions. This constrains the input/output sizes of the whole network. Such constraint may not be problematic for image classification problems \cite{simonyan2014very}. However, it can limit the applicability of rendering techniques for two main reasons: 1) The rendering quality is directly related to the spatial resolution 2) Images commonly come in different sizes \cite{kim2015accurate}. Compared to Deep3D, our work takes various-sized images as input and outputs correspondingly-sized images in a single network. This is due our use of FCN  without any dense layer or pretrained models. 

\section{Proposed method}

\begin{figure}[!t]
\begin{center}
\epsfig{file=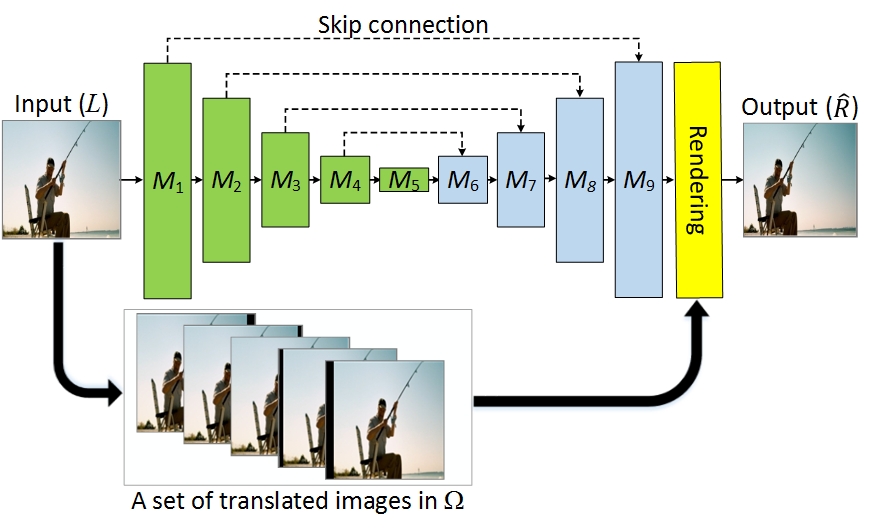,width=8cm}
\caption{DeepView$_{gen}$ architecture. The encoding, decoding and rendering networks are shown in green, blue and yellow respectively. The encoding network extracts low, middle and high level features from the input image and transfers them to the decoding network. After decoding, the rendering network generates probabilistic disparity maps and estimates the right-image. Here, a set of translated images are used (see Fig. 2).}
\end{center}
\vspace{-0.4cm}
\end{figure}

\vspace{0.4cm}
\noindent
{\bf FCN with rendering network}
Fig. 1 illustrates our DeepView$_{ren}$ architecture. The green, blue and yellow colors define the encoding, decoding, and rendering networks, respectively. Conceptually, the encoding network extracts low, middle and high level features from the input image and transfers them to the decoding network. After decoding, the rendering network generates probabilistic disparity maps. The right-image $\hat{R}$ is rendered using the set of images translated by the disparity range $\Omega$. In this paper, we establish an FCN architecture based on units of convolution modules ($M$) each of which consists of a set of $K$ convolution layers followed by an activation unit. We use rectification linear units (ReLU) for activation. The encoding and decoding networks comprises total 9 modules ($M_{i}$, $i = 1, ..., 9$) having a total of $9 \cdot K$ convolution layers in our architecture. For simplified description, we denote the $j$-th convolution layer in the $i$-th convolution module as Conv$_{i,j}$. Note that ReLU is not applied for the last convolution layer of the decoding network as the rendering network contains a softmax activation unit. This  normalizes the output of the decoding network.

Estimating depth requires wide contextual information from the entire scene \cite{eigen2014depth}. In order to secure large receptive fields for our network, we propose to use multiple down-scaling and up-scaling convolution (deconvolution) layers with strides of 2 and 0.5, respectively. That is, the first convolution layers of $M_{i}, i =1,...,4$ convolve with a stride of 2, and the last convolution layers of $M_{i}, i =6,...,9$ convolve with a stride of 0.5. As in \cite{mao2016image, he2016deep}, we adopted the skip connections with additions which transfer sharp object details to the decoding network. That is, the outputs of Conv$_{1,K-1}$, Conv$_{2,K-1}$, Conv$_{3,K-1}$, and Conv$_{4,K-1}$ in the encoding network are connected to the inputs of Conv$_{6,2}$, Conv$_{7,2}$, Conv$_{8,2}$, and Conv$_{9,2}$ in the decoding network, respectively.  

\begin{figure}[!t]
\begin{center}
\epsfig{file=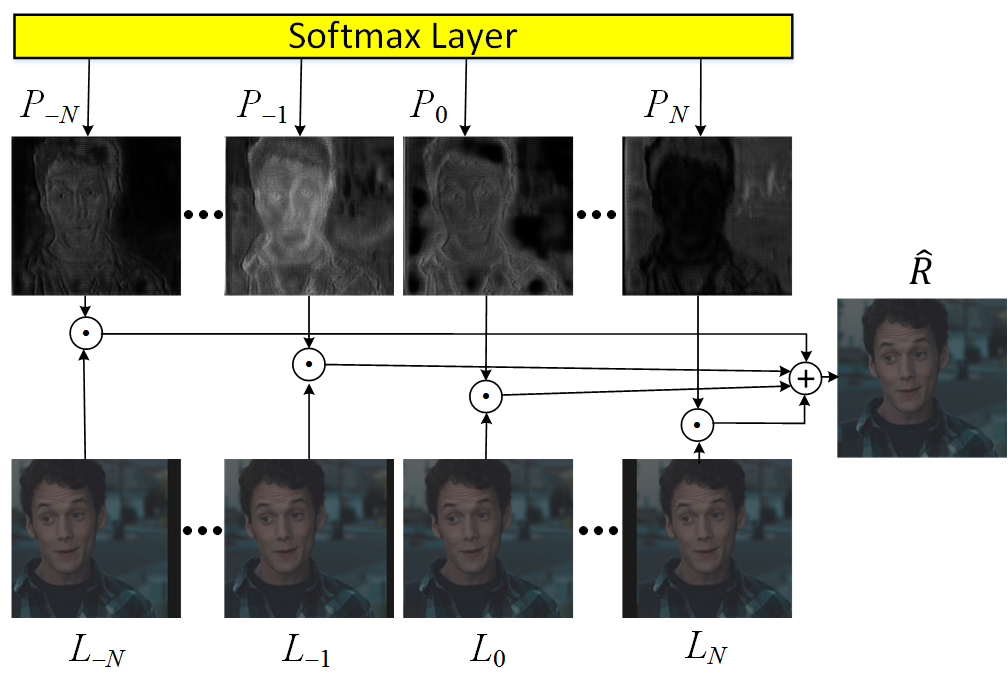,width=8cm}
\caption{The rendering network. The softmax layer normalizes the output of the decoding network to probabilistic values over channels ($P_{i}, i \in \Omega$). Here, the number of channels is identical to the number of values in a disparity range $\Omega=\left \{  -N,-N+1,...,0,1,...,N \right \}$. The final right-view image $\hat{R}$ is synthesized by pixel-wise multiplication between $P$ and their correspondingly-translated left-images $L$. }
\end{center}
\vspace{-0.4cm}
\end{figure}

Fig. 2 shows the rendering network. The softmax layer normalizes the output values of the decoding network to be probabilistic values over channels ($P_{i}, i \in \Omega$), wheres the number of channels is identical to the number of values in a disparity range $\Omega=\left \{  -N,-N+1,...,1,0,...,N \right \}$. Since the softmax layer approximates the max operation, it gives sparsity in selecting disparity values at each pixel location. The final right-view image $\hat{R}$ is synthesized by pixel-wise multiplication between $P$ and their correspondingly-translated left-images $L$, which can mathematically be expressed as 
\be
\hat{R}(x,y,c)=\sum_{i\in \Omega }P_{i}(x,y)L_{i}(x,y,c)
\label{eq_3}
\ee
where $(x,y) \in H \times W $ is the pixel index in a $H \times W$-sized image, and $c$ is the index for RGB color channels. 



\vspace{0.4cm}
\noindent
{\bf Decoupled network}
Fig. 3 illustrates our decoupled structure DeepView$_{dec}$. This structure processes the chrominance and luminance channels separately. It consists of two decoupled networks having the same architecture, \ie, luminance (Y) and chrominance (Cb, Cr). Each network trains and infers separately. The green, blue colors in each network define the encoding and decoding networks, respectively, while the yellow color indicates the color conversion block between RGB and YCbCr. The RGB-image-input is converted to Y, Cb, and Cr images where Cb and Cr images are further downscaled with factor of 2. Conceptually, the encoding network extracts low, middle and high level features from the input image and transfers them to the decoding network. The inferred images by the decoding network are inverted to the output RGB image. Note that the luminance network is trained with only Y channel images while chrominance network is trained with both Cb and Cr channel images. 

\begin{figure}[!t]
\begin{center}
\epsfig{file=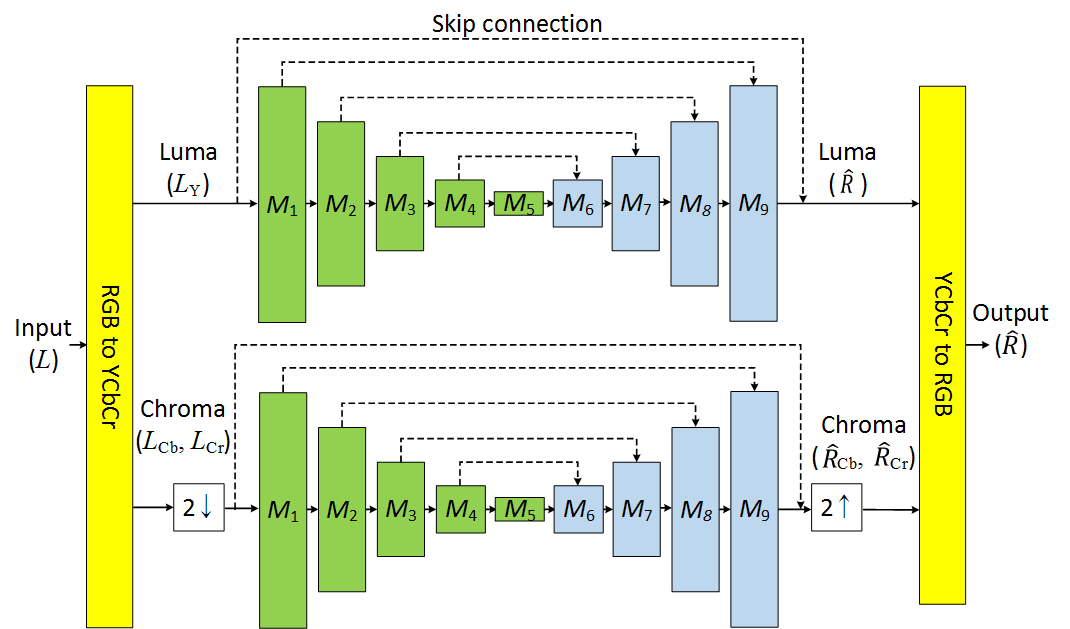,width=8cm}
\caption{DeepView$_{dec}$ architecture. DeepView$_{dec}$ consists of two decoupled networks having the same architecture, \ie, luminance (Y) and chrominance (Cb, Cr) network. Each network is trained separately.}
\end{center}
\vspace{-0.4cm}
\end{figure}

\section{New dataset for SIVG}
There are some large RGB-depth (or relative depth) databases, such as KITTI\cite{geiger2012we}, NYU \cite{silberman2012indoor} and DIW \cite{chen2016single}. Such datasets have effectively been used for training and testing depth-estimation methods. To the best of our knowledge, there are no publicly available large datasets of stereoscopic image pairs. In this paper, we introduce a new large dataset for SIVG. Our dataset is collected from 27 non-animated stereo movies having Full-HD (1920$\times$1080) resolutions. Fig. 4 shows some thumbnail-images of our dataset containing a variety of genres including action, adventure, drama, fantasy, romance, horror, etc. For generating the dataset, we eliminated the text-only frames at the beginning and ending of the movies. The final valid frames have total 42.5 hours duration with 2M frames. We will publicly release our dataset for research purpose.

\begin{figure}[!t]
\begin{center}
\epsfig{file=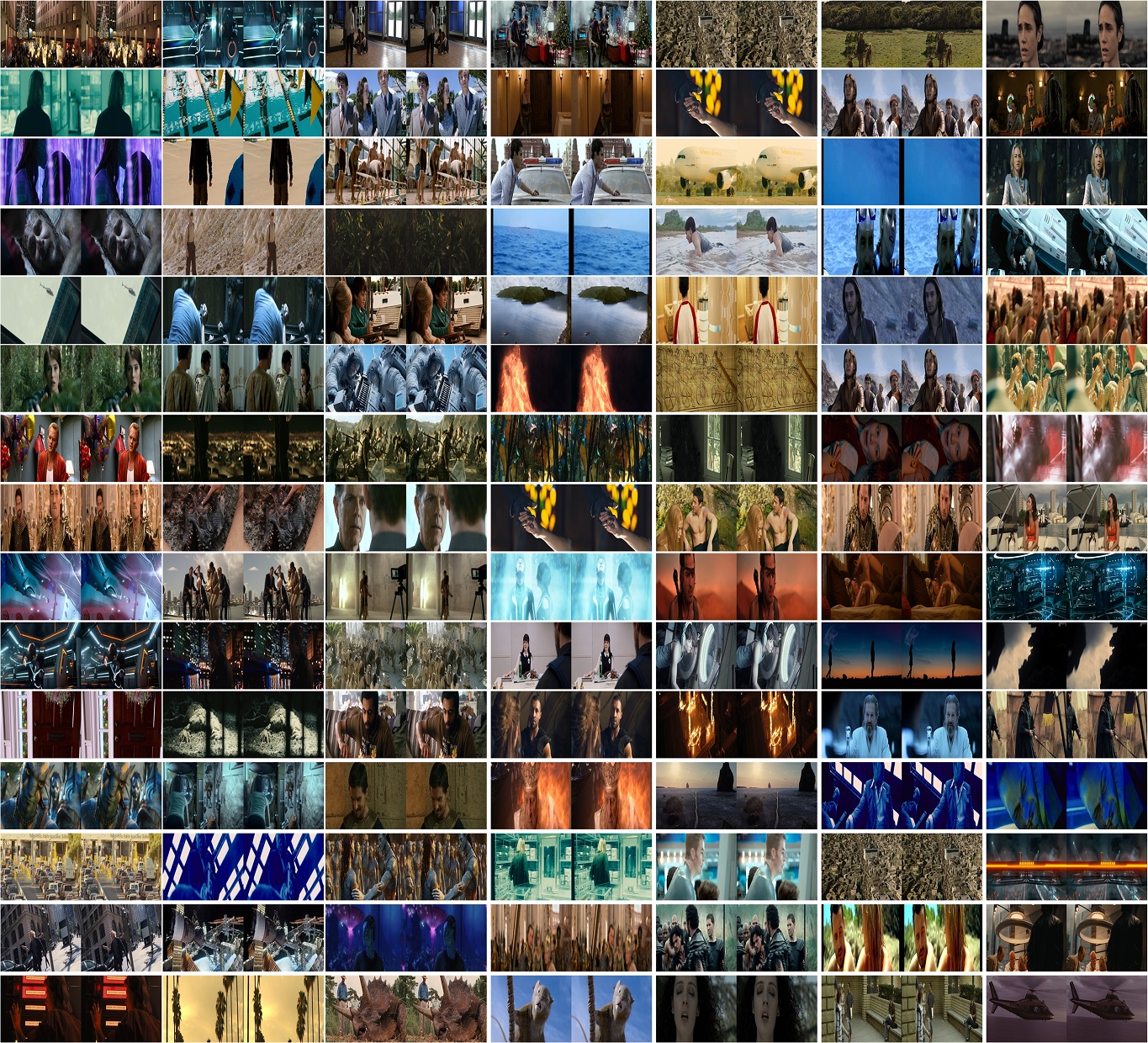,width=8cm}
\caption{Some thumbnail-images of our dataset of  2M stereoscopic image pairs.}
\end{center}
\vspace{-0.4cm}
\end{figure}

\section{Deep analysis of the proposed method}
We perform comprehensive experiments to explore optimal network architectures for SIVG in terms of prediction accuracy and computational efficiency. We also explore spatial scalability of our method.  

\subsection{Implementation details}
Our architectures are implemented based on MatConvNet, a Matlab-based CNN library \cite{vedaldi2015matconvnet}. During training, we minimize the mean squared error (MSE) over training data, \ie, MSE = $(Z^{-1})||\hat{R}-R ||^{2}_{F}$, where $Z$ is the number of pixels in an image, and $\left \| \cdot  \right \|_{F}$ is the Frobenius norm. 

Regarding the number of convolution layers, we set $K=4$ for each module ($M$) as a default setting by considering trade-off between accuracy and efficiency. This leads to total 36 convolution layers in our architectures. We set the filter size of each convolutional layer to $3\times3\times64\times64$ $ \in \mathbb{R}^{H\times W \times D \times C}$ (for deconvolution layers, we set their filter sizes to $4\times4\times64\times64$), where $C$ is the number of filters in a convolution layer, and $H$, $W$ and $D$ correspond to the height, width and depth of each filter.
 
For the last convolution layer in the decoding network of DeepView$_{ren}$, we set its filter size to $3\times3\times64\times33$ as we set disparity range $\Omega = \left \{-15,-14,...,16, 17 \right \}$. For DeepView$_{dec}$, the number of filter in the last convolution layer is 1 as its input and output are Y or CbCr. 

To optimize the network, we use Adam solver with $\beta_{1}= 0.9$, $\beta_{2}= 0.9999$, $\epsilon = 10^{-8}$ \cite{kingma2014adam}. We set the training-batch-size to 64. To initialize the weights in convolution layers, we followed the method of \cite{he2015delving}. We trained our networks with a total of 30 epochs with the fixed learning rate of $10^{-4}$. It takes one day for training in a single Nvidia GTX Titan X GPU with 12GB memory. The aforementioned training configurations are identically used in all experiments unless otherwise mentioned.

Since there are no publicly available SIVG datasets, we use our dataset introduced in Section 4. Our dataset consists of 27 non-animated stereoscopic movies. We divided them into 18 training and 9 testing movies, such that there is no overlap between the training and testing datasets. To reduce computational complexity for training, we selected training/testing frames every 2 seconds among a total of 2M frames, resulting in 58K training frames and 22K testing frames. For all the training/testing frames, we performed a downscaling process by preserving the frame aspect ratio and slightly cropped the upper and lower pixel boundaries, such that the spatial resolution of all the training/testing frames becomes 384$\times$160. 
 
To measure computational efficiency, we use memory consumption (\#Param), \ie, the number of weight and bias values used in convolution and batch-normalization layers. Those parameters should be kept in the memory during the entire process. Also, the average running speed in frames per second (\textit{fps}) is measured for all the testing images. We use MSE and mean absolute error (MAE) to measure prediction accuracy. MAE is calculated as $E[|\hat{R}-R|]$, where $E[\cdot]$ is expectation operator over pixels in an image. Note that, higher $fps$ indicates higher performance, while higher \#Param, MSE and MAE mean lower performance.

\subsection{Effectiveness of rendering network}
We verify the effectiveness of the rendering network in DeepView$_{ren}$ by performing experiments on DeepView$_{ren}$ with and without the rendering network. Table 1 shows the performance of DeepView$_{ren}$ with and without the rendering network. As shown in Table 1, the rendering network improves prediction accuracy in terms of MSE and MAE. It also does not introduce noticeable computational complexity in both $fps$ and \#Param. This is because the rendering network explicitly performs pixel-translations of the left-image and selects the best translation based on the generated disparity maps. 


\begin{table}[!t]
\begin{center}
\begin{tabular}{crrrr}
\hline
Method  & 		without RN  & 	withRN      \\
\hline\hline
$fps$	&	52.19   &	51.18    \\
\#Param	&	1.40M	&	1.40M    \\ \hline
MSE	    &	218.04	&	213.04    \\
MAE 	&	5.77	&	5.54    \\
\hline
\end{tabular}
\end{center}
\caption{Performance of DeepView$_{ren}$ with/without the rendering network (RN).}
\end{table}

\subsection{Spatial scalability} 
Our architectures have spatial scalability, \ie, they can support multiple input/output sizes in a single network. To verify the spatial scalability, we train DeepView$_{rec}$ for various-sized images. In order to generate the training/testing with different spatial resolutions, we use the same 18 and 9 movies for training and testing, respectively, as described in Section 5.1. However, here we downscaled the data with different factors of $(4,5,6)$. This generates 3 datasets of the same content but at different scales. We use this data to train our DeepView$_{rec}$. We performed four trainings. The first three train each scale separately while the last trains all scales together at once. The performance of each model is calculated. An architecture is scalable if the testing performance of all the all scales model is similar to the performance of the scale-specific model. 

Table 2 shows the prediction performance of DeepView$_{rec}$ for different scale training/testing datasets. As shown in Table 2, the dedicated model for each scale shows the lowest error in correspondingly-sized testing data. While the performance of the trained model with all the scales also approximates all the testing data. This indicates that our approach can support spatial scalability in a single network. Fig. 4 shows the qualitative performance of DeepView$_{rec}$ for three different spatial resolutions (320$\times$128, 384$\times$160, and 480$\times$192). In Fig. 4, the upper and bottom rows illustrate the estimated right-view images and their ground-truth, respectively. As shown in Fig. 4, DeepView$_{rec}$ is able to estimate right-view images for three different spatial resolutions consistently in a single network. Therefore, contrary to the existing method requiring dedicated networks depending on different spatial scales, our architectures can effectively be used for practical applications in a single network.


\begin{figure*}[!t]
\begin{center}
\epsfig{file=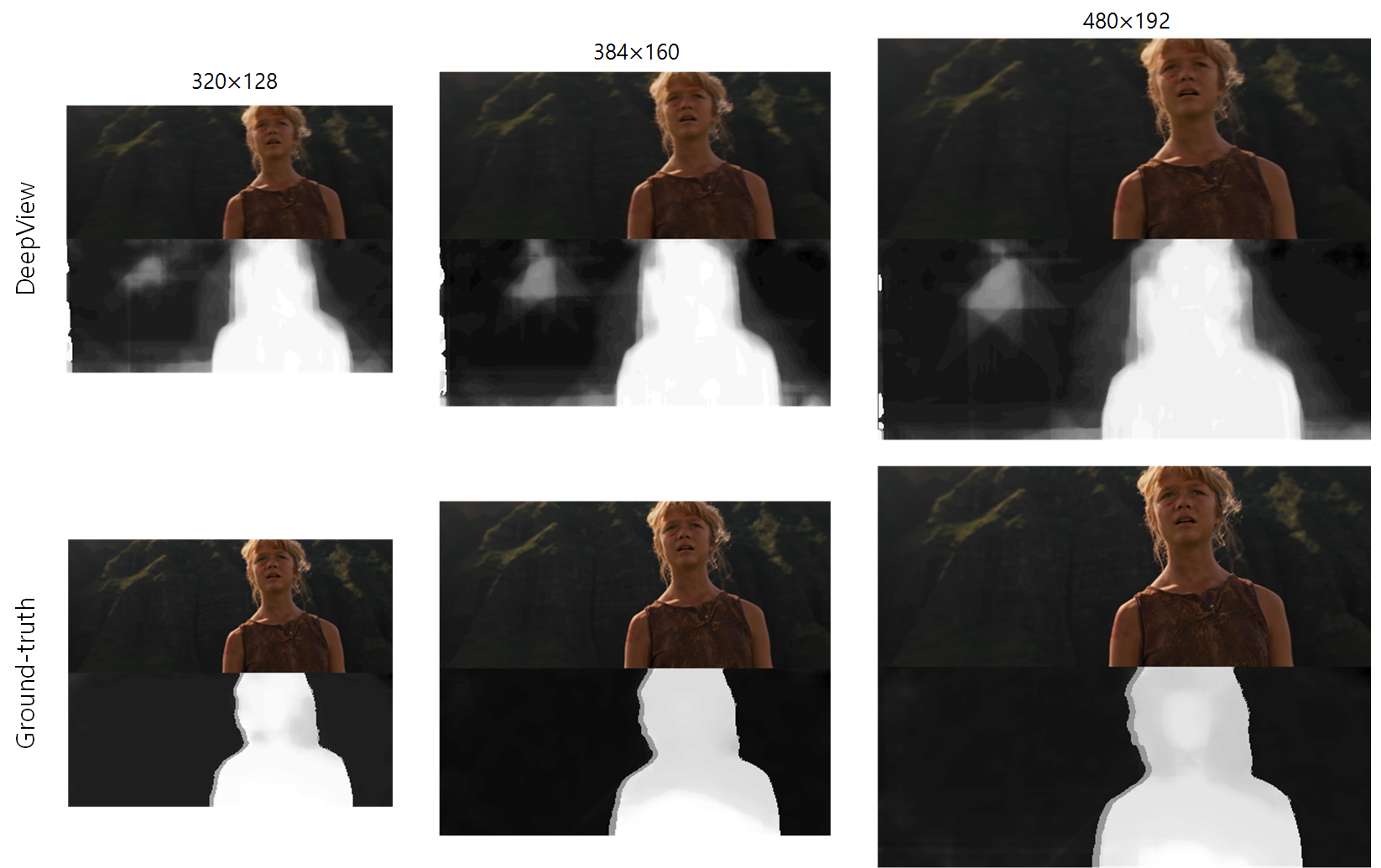,width=15cm}
\caption{Qualitative performance of DeepView$_{rec}$ on three different spatial resolutions (320$\times$128, 384$\times$160, and 480$\times$192).}
\end{center}
\vspace{-0.4cm}
\end{figure*}


\begin{table}[!t]
\begin{center}
\label{my-label}
\begin{tabular}{crrrrr}
\hline
\multicolumn{2}{l}{Test$\setminus$Train} & Scale$_{4}$ & Scale$_{5}$ & Scale$_{6}$ & Scale$_{4,5,6}$ \\
\hline\hline
\multirow{2}{*}{Scale$_{4}$}    & MSE  & 216.18     & 218.79         & 220.48     & 218.79  \\
                                & MAE  & 5.67       & 5.70           & 5.79       & 5.70  \\
\hline
\multirow{2}{*}{Scale$_{5}$}    & MSE  & 217.49      & 213.04       & 219.49     & 217.86  \\
                                & MAE  & 5.74      & 5.54      & 5.75     & 5.66  \\
\hline
\multirow{2}{*}{Scale$_{6}$}    & MSE  & 197.20      & 194.12         & 193.58     & 195.13  \\
                                & MAE  & 5.68      & 5.44         & 5.34     & 5.47  \\
\hline
\end{tabular}
\end{center}
\caption{Prediction performance of DeepView$_{rec}$ for different scale training/testing datasets.}
\end{table}

\section{Experiments}
To verify the effectiveness of our architectures, we compare them against the state of the art Deep3D \cite{xie2016deep3d}. Note that we do not compare our approach with the existing depth-estimation methods, since they require an additional DIBR process to generate right-images. This often involves much computation and visual distortion \cite{xie2016deep3d}. It is shown in \cite{xie2016deep3d} that Deep3D remarkably outperforms the existing depth-estimation method followed by a DIBR.

\subsection{Objective performance}
In the objective performance evaluation, we compare the estimated right-images with their ground-truth right-images. We also report the baseline performance that is measured with the ground-truth left- and right-images, (\ie, ground-truth left-images are considered estimated right-images in baseline). Table 3 shows the prediction performance of baseline, Deep3D, DeepView$_{ren}$ and DeepView$_{dec}$. The best performance in each column is highlighted in black bold. As shown in Table 3, Deep$_{ren}$ shows competitive performance to Deep3D while DeepView$_{dec}$ outperforms both baseline and Deep3D for both MSE and MAE.

\begin{table}
\begin{center}
\begin{tabular}{crr}
\hline
Method                  & 		MSE              &	MAE         \\ 
\hline\hline				
Base	                &	259.54	              &	6.23        \\ 
Deep3D	                &	213.40	              &	5.72        \\ 
DeepView$_{ren}$	    &	213.04	             &	5.54        \\ 
DeepView$_{dec}$	    &	\textbf{190.27}	    &	\textbf{5.46}   \\ 
\hline
\end{tabular}
\end{center}
\caption{Prediction performance of baseline (Base), Deep3D, DeepView$_{ren}$ and DeepView$_{dec}$. The best performance in each column is highlighted in black bold.}
\end{table}

Fig. 6 shows the qualitative performance of baseline, Deep3D and DeepView$_{dec}$. In Fig. 6, the first, second and third columns are results of Deep3D, DeepView$_{dec}$ and ground-truth, respectively. Each row in Fig. 6 consists of the estimated right-view image and depth map measured with original left-view and estimated right-view image pairs. The disparity maps are estimated by using a block-based stereo-matching method \cite{hirschmuller2005accurate}. Note that our method does not aim at estimating accurate disparity maps. The disparity maps in Fig. 6 are illustrated only to show the consistency between the estimated right-view images and their ground-truth in depth perception. As shown in Fig. 6, the proposed DeepView$_{dec}$ produces sharper edges (in the red and yellow boxes) and depth consistency with the ground-truth compared to Deep3D.

\begin{figure*}[!t]
\begin{center}
\epsfig{file=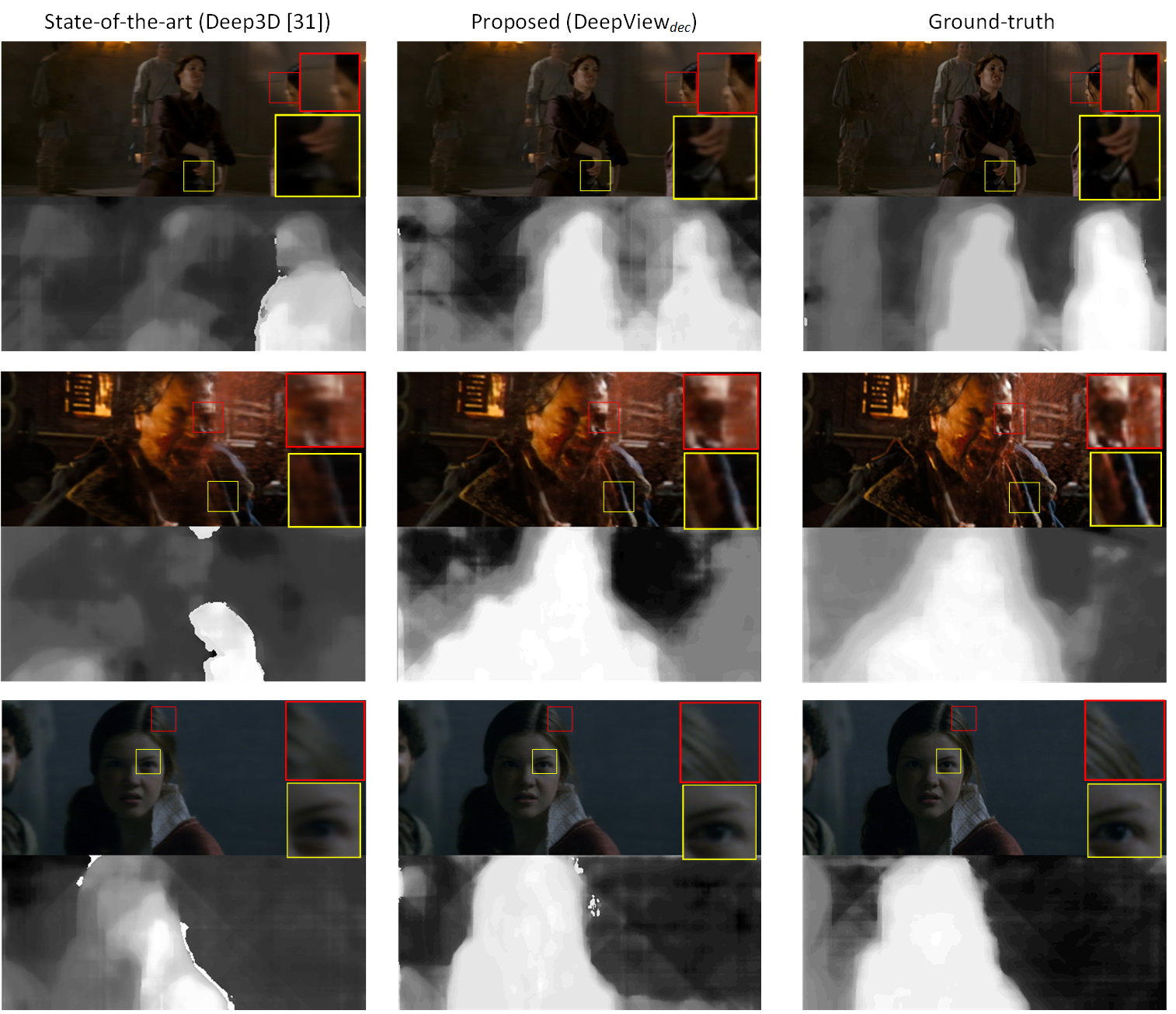,width=16cm}
\caption{Qualitative performance comparison of Deep3D and DeepView$_{dec}$ with ground-truth. The red and yellow boxes show that our DeepView$_{dec}$ tends to produce sharper edges compared to Deep3D.}
\end{center}
\vspace{-0.4cm}
\end{figure*}

\subsection{Subjective performance evaluation} 
We perform subjective quality assessment experiments to verify the effectiveness of DeepView$_{dec}$. For this, we use stereoscopic images made with the pairs of original left-view images and estimated right-view images. Table 3 summarizes the experimental setup for our subjective experiments. 

\begin{table}
\begin{center}
 \begin{adjustbox}{max width=\textwidth}
\begin{tabular}{l||l}
\hline
Display	& ZM-M240W, Polarized 3D\\ 
			& (24 inch, 1920$\times$1080 Full-HD) \\ \hline

\# Subject 	& 15 (12 male, 3 female) \\ 
 					& (mean age: 29.8) \\  \hline

Viewing distance 		& 0.5m \\ \hline
Ambient illum.	& 200 lux \\ \hline 
Testing image size		& 384$\times$160\\
\hline
\end{tabular}
\end{adjustbox}
\end{center}
\caption{Experimental setup for subjective quality assessment.}
\end{table}

We randomly select 100 image pairs from the testing dataset and used the adjectival categorical judgment method  \cite{recommendation2002500} where the reference ground-truth stereoscopic images (made with original left- and right-view images) and the compared stereoscopic images (made with original left- and estimated right-view images) are vertically juxtaposed with pseudo-random order. In the adjectival categorical judgment method \cite{recommendation2002500}, the subjects evaluate their perceived qualities of the presented images being compared. The comparison scale for the comparison images is given with -3 as `Much worse', -2 as `Worse', -1 as `Slightly worse', 0 as `The same', +1 as `Slightly better', +2 as `Better', +3 as `Much Better' against their reference images. Note that the negative scales imply that the compared stereoscopic images are perceived worse than ground-truth ones. The individual comparison scores are provided in average as mean opinion score (MOS). As a result, the MOS values of the proposed DeepView$_{dec}$ and Deep3D were $-0.37$ and $-0.48$, respectively. This indicates that DeepView$_{dec}$ produces better visual quality compared to the state-of-the-art, Deep3D.

\subsection{Computation efficiency} The memory consumption in \#Param. and computation speed in $fps$ between Deep3D and DeepView$_{dec}$ are compared. Note that our DeepView$_{dec}$ is implemented on Matlab with MatConvNet library while Deep3D is implemented on Python with MXNet \cite{chen2015mxnet}. For a comparison, we implemented the same architecture of Deep3D on Matlab with MatConvnet and measured the running speed. Table. 5 compares  Deep3D and DeepView$_{dec}$ in terms of memory consumption in \#Param and running speed in $fps$. As shown in Table. 5, DeepView$_{dec}$ runs  5.1 times faster with 24 times lower memory consumption. Note that the heavy computation and memory consumption of Deep3D comes mostly from the high-level convolution layers and the dense layers in the pre-trained CNN. Those layers help the network capturing global contextual information in a whole image. Contrary to Deep3D, we use multiple up- and down-scale convolution layers with symmetric encoding/decoding networks to secure large receptive field sizes. Most convolution layers in our network have 3$\times$3$\times$64$\times$64 filter sizes, requiring relatively much lower computation complexity and memory consumption compared to the convolution and dense layers used in Deep3D.

\subsection{Limitation}
 We found out that our trained network is slightly overfitted, i.e., the gap between validation loss and the training loss are not negligible. This indicates that the proposed method can be further be improved by using more a mount of data or effective data augmentation methods.

\begin{table}[!t]
\begin{center}
\begin{tabular}{crrrr}
\hline
Method  & 		Deep3D  & 	DeepView$_{rec}$    &	DeepView$_{dec}$      \\
\hline\hline
$fps$	&	9.57   &	52.19   &	23.92   \\
\#Param	&	33.52M	&	1.40M   &	2.80M   \\ \hline
\end{tabular}
\end{center}
\caption{Comparison of Deep3D and DeepView$_{dec}$ in terms of \#Param and \textit{fps}. }
\end{table}

\section{Conclusion}
We proposed the use of fully convolutional networks for the problem of novel view synthesis from single images. Our solution directly learns the transfer from the left input image to the right image, without explicit estimation of depth maps. We presented two architectures with the aim to reduce prediction error as well as the computational complexity and memory consumption. One network makes use of a rendering network while the other is based on separated decoupled processing for the chrominance and luminance channels. The former network achieves competitive performance however with significantly less computational and memory consumption (x5 times faster speed with x24 times lower memory consumption). The decoupled structure is slightly more expensive, but significantly less prediction error than the state of the art. We also presented a large dataset of stereoscopic movies suitable for training such networks. We examined our network through objective and subjective measures. Future work can address utilizing other types of input data (\eg, depth and segmentation) for better performance.


{\small
\begin{thebibliography}{10}\itemsep=-1pt
	
	\bibitem{appia2014fully}
	V.~Appia and U.~Batur.
	\newblock Fully automatic 2d to 3d conversion with aid of high-level image
	features.
	\newblock In {\em IS\&T/SPIE Electronic Imaging}, pages 90110W--90110W.
	International Society for Optics and Photonics, 2014.
	
	\bibitem{avila2014virtual}
	L.~Avila and M.~Bailey.
	\newblock Virtual reality for the masses.
	\newblock {\em IEEE computer graphics and applications}, 34(5):103--104, 2014.
	
	\bibitem{baig2014im2depth}
	M.~H. Baig, V.~Jagadeesh, R.~Piramuthu, A.~Bhardwaj, W.~Di, and N.~Sundaresan.
	\newblock Im2depth: Scalable exemplar based depth transfer.
	\newblock In {\em IEEE Winter Conference on Applications of Computer Vision},
	pages 145--152. IEEE, 2014.
	
	\bibitem{calagari2015gradient}
	K.~Calagari, M.~Elgharib, P.~Didyk, A.~Kaspar, W.~Matusik, and M.~Hefeeda.
	\newblock Gradient-based 2d-to-3d conversion for soccer videos.
	\newblock In {\em Proceedings of the 23rd ACM International Conference on
		Multimedia}, pages 331--340. ACM, 2015.
	
	\bibitem{chen2015mxnet}
	T.~Chen, M.~Li, Y.~Li, M.~Lin, N.~Wang, M.~Wang, T.~Xiao, B.~Xu, C.~Zhang, and
	Z.~Zhang.
	\newblock Mxnet: A flexible and efficient machine learning library for
	heterogeneous distributed systems.
	\newblock {\em arXiv preprint arXiv:1512.01274}, 2015.
	
	\bibitem{chen2016single}
	W.~Chen, Z.~Fu, D.~Yang, and J.~Deng.
	\newblock Single-image depth perception in the wild.
	\newblock In {\em Advances in Neural Information Processing Systems}, pages
	730--738, 2016.
	
	\bibitem{cozman1997depth}
	F.~Cozman and E.~Krotkov.
	\newblock Depth from scattering.
	\newblock In {\em Computer Vision and Pattern Recognition, 1997. Proceedings.,
		1997 IEEE Computer Society Conference on}, pages 801--806. IEEE, 1997.
	
	\bibitem{dong2016image}
	C.~Dong, C.~C. Loy, K.~He, and X.~Tang.
	\newblock Image super-resolution using deep convolutional networks.
	\newblock {\em IEEE Transactions on Pattern Analysis and Machine Intelligence},
	38(2):295--307, 2016.
	
	\bibitem{eigen2014depth}
	D.~Eigen, C.~Puhrsch, and R.~Fergus.
	\newblock Depth map prediction from a single image using a multi-scale deep
	network.
	\newblock In {\em Advances in Neural Information Processing Systems}, pages
	2366--2374, 2014.
	
	\bibitem{flynn2015deepstereo}
	J.~Flynn, I.~Neulander, J.~Philbin, and N.~Snavely.
	\newblock Deepstereo: Learning to predict new views from the world's imagery.
	\newblock {\em arXiv preprint arXiv:1506.06825}, 2015.
	
	\bibitem{geiger2012we}
	A.~Geiger, P.~Lenz, and R.~Urtasun.
	\newblock Are we ready for autonomous driving? the kitti vision benchmark
	suite.
	\newblock In {\em Computer Vision and Pattern Recognition (CVPR), 2012 IEEE
		Conference on}, pages 3354--3361. IEEE, 2012.
	
	\bibitem{he2015delving}
	K.~He, X.~Zhang, S.~Ren, and J.~Sun.
	\newblock Delving deep into rectifiers: Surpassing human-level performance on
	imagenet classification.
	\newblock In {\em Proceedings of the IEEE International Conference on Computer
		Vision}, pages 1026--1034, 2015.
	
	\bibitem{he2016deep}
	K.~He, X.~Zhang, S.~Ren, and J.~Sun.
	\newblock Deep residual learning for image recognition.
	\newblock In {\em Proceedings of the IEEE Conference on Computer Vision and
		Pattern Recognition}, pages 770--778, 2016.
	
	\bibitem{hirschmuller2005accurate}
	H.~Hirschmuller.
	\newblock Accurate and efficient stereo processing by semi-global matching and
	mutual information.
	\newblock In {\em 2005 IEEE Computer Society Conference on Computer Vision and
		Pattern Recognition (CVPR'05)}, volume~2, pages 807--814. IEEE, 2005.
	
	\bibitem{hoiem2005automatic}
	D.~Hoiem, A.~A. Efros, and M.~Hebert.
	\newblock Automatic photo pop-up.
	\newblock {\em ACM Transactions on Graphics (TOG)}, 24(3):577--584, 2005.
	
	\bibitem{kalantari2016learning}
	N.~K. Kalantari, T.-C. Wang, and R.~Ramamoorthi.
	\newblock Learning-based view synthesis for light field cameras.
	\newblock {\em arXiv preprint arXiv:1609.02974}, 2016.
	
	\bibitem{karsch2014depth}
	K.~Karsch, C.~Liu, and S.~B. Kang.
	\newblock Depth transfer: Depth extraction from video using non-parametric
	sampling.
	\newblock {\em IEEE Transactions on Pattern Analysis and Machine Intelligence},
	36(11):2144--2158, 2014.
	
	\bibitem{kim20102d}
	J.~Kim, A.~Baik, Y.~J. Jung, and D.~Park.
	\newblock 2d-to-3d conversion by using visual attention analysis.
	\newblock In {\em IS\&T/SPIE Electronic Imaging}, pages 752412--752412.
	International Society for Optics and Photonics, 2010.
	
	\bibitem{kim2015accurate}
	J.~Kim, J.~K. Lee, and K.~M. Lee.
	\newblock Accurate image super-resolution using very deep convolutional
	networks.
	\newblock {\em arXiv preprint arXiv:1511.04587}, 2015.
	
	\bibitem{kim2015deeply}
	J.~Kim, J.~K. Lee, and K.~M. Lee.
	\newblock Deeply-recursive convolutional network for image super-resolution.
	\newblock {\em arXiv preprint arXiv:1511.04491}, 2015.
	
	\bibitem{kingma2014adam}
	D.~Kingma and J.~Ba.
	\newblock Adam: A method for stochastic optimization.
	\newblock {\em arXiv preprint arXiv:1412.6980}, 2014.
	
	\bibitem{konrad2013learning}
	J.~Konrad, M.~Wang, P.~Ishwar, C.~Wu, and D.~Mukherjee.
	\newblock Learning-based, automatic 2d-to-3d image and video conversion.
	\newblock {\em IEEE Transactions on Image Processing}, 22(9):3485--3496, 2013.
	
	\bibitem{liu2015deep}
	F.~Liu, C.~Shen, and G.~Lin.
	\newblock Deep convolutional neural fields for depth estimation from a single
	image.
	\newblock In {\em Proceedings of the IEEE Conference on Computer Vision and
		Pattern Recognition}, pages 5162--5170, 2015.
	
	\bibitem{long2015fully}
	J.~Long, E.~Shelhamer, and T.~Darrell.
	\newblock Fully convolutional networks for semantic segmentation.
	\newblock In {\em Proceedings of the IEEE Conference on Computer Vision and
		Pattern Recognition}, pages 3431--3440, 2015.
	
	\bibitem{mao2016image}
	X.~Mao, C.~Shen, and Y.-B. Yang.
	\newblock Image restoration using very deep convolutional encoder-decoder
	networks with symmetric skip connections.
	\newblock In {\em Advances in Neural Information Processing Systems}, pages
	2802--2810, 2016.
	
	\bibitem{matusik20043d}
	W.~Matusik and H.~Pfister.
	\newblock 3d tv: a scalable system for real-time acquisition, transmission, and
	autostereoscopic display of dynamic scenes.
	\newblock {\em ACM Transactions on Graphics (TOG)}, 23(3):814--824, 2004.
	
	\bibitem{recommendation2002500}
	I.~Recommendation.
	\newblock 500-11,?쐌ethodology for the subjective assessment of the quality of
	television pictures,??recommendation itu-r bt. 500-11.
	\newblock {\em ITU Telecom. Standardization Sector of ITU}, 2002.
	
	\bibitem{saxena2009make3d}
	A.~Saxena, M.~Sun, and A.~Y. Ng.
	\newblock Make3d: Learning 3d scene structure from a single still image.
	\newblock {\em IEEE Transactions on Pattern Analysis and Machine Intelligence},
	31(5):824--840, 2009.
	
	\bibitem{silberman2012indoor}
	N.~Silberman, D.~Hoiem, P.~Kohli, and R.~Fergus.
	\newblock Indoor segmentation and support inference from rgbd images.
	\newblock In {\em European Conference on Computer Vision}, pages 746--760.
	Springer, 2012.
	
	\bibitem{simonyan2014very}
	K.~Simonyan and A.~Zisserman.
	\newblock Very deep convolutional networks for large-scale image recognition.
	\newblock {\em arXiv preprint arXiv:1409.1556}, 2014.
	
	\bibitem{vedaldi2015matconvnet}
	A.~Vedaldi and K.~Lenc.
	\newblock Matconvnet: Convolutional neural networks for matlab.
	\newblock In {\em Proceedings of the 23rd ACM international conference on
		Multimedia}, pages 689--692. ACM, 2015.
	
	\bibitem{xie2016deep3d}
	J.~Xie, R.~Girshick, and A.~Farhadi.
	\newblock Deep3d: Fully automatic 2d-to-3d video conversion with deep
	convolutional neural networks.
	\newblock {\em arXiv preprint arXiv:1604.03650}, 2016.
	
	\bibitem{zhuo2009recovery}
	S.~Zhuo and T.~Sim.
	\newblock On the recovery of depth from a single defocused image.
	\newblock In {\em International Conference on Computer Analysis of Images and
		Patterns}, pages 889--897. Springer, 2009.
	
\end{thebibliography}

}
\end{document}